\documentclass[10pt,twocolumn,letterpaper]{article}

\usepackage{cvpr}              
\usepackage[accsupp]{axessibility}

\usepackage{algorithm}
\usepackage{algpseudocode} 
\usepackage{amsmath}

\newcommand{\M}[1]{\mathtt{#1}}
\newcommand{\V}[1]{\mathbf{#1}}
\definecolor{cvprblue}{rgb}{0.21,0.49,0.74}
\usepackage[pagebackref,breaklinks,colorlinks,allcolors=cvprblue]{hyperref}

\def\paperID{6688} 
\def\confName{CVPR}
\def\confYear{2026}

\title{Solving Minimal Problems Without Matrix Inversion Using FFT-Based Interpolation}

\author{
Haidong Wu\hspace{1cm}Snehal Bhayani\hspace{1cm}Janne Heikkilä\\
Center for Machine Vision and Signal Analysis\\
University of Oulu, Oulu, Finland\\
{\tt\small Haidong.Wu@oulu.fi, \tt\small Snehal.Bhayani@oulu.fi, \tt\small Janne.Heikkila@oulu.fi}}

\begin{document}
\maketitle
\begin{abstract}
Estimating camera geometry typically involves solving minimal problems formulated as systems of multivariate polynomial equations, which often pose computational challenges when using existing Gröbner-basis or resultant-based methods due to matrix inversion needed in the online solver. Here we propose a sampling-based, matrix inversion-free method that constructs the solvers using sparse hidden-variable resultants. The determinant polynomial in the hidden variable is efficiently reconstructed via inverse fast Fourier transform interpolation from sampled evaluations, avoiding symbolic expansion. Solving this polynomial yields the hidden variable, and the remaining unknowns are recovered by identifying rank-1 deficient submatrices and applying Cramer's rule. A greatest common divisor-based criterion ensures robust submatrix identification under noise. Experiments on diverse minimal problems demonstrate that the proposed solver achieves strong numerical stability and competitive runtime, particularly for small-scale problems, providing a practical alternative to traditional Gröbner-basis and resultant-based solvers.

\end{abstract}    
\section{Introduction}
\label{sec:intro}
Estimating camera geometry is a fundamental problem in computer vision, with applications in structure-from-motion~\cite{snavely2008modeling}, visual navigation~\cite{scaramuzza2011visual}, large-scale 3D reconstruction~\cite{jared2015reconstructing}, and image localization~\cite{sattler2016efficient}.
In practice, such estimation often relies on noisy measurements or sparse correspondences, which necessitates the use of robust methods. A widely adopted strategy is to solve so-called minimal problems~\cite{kukelova2013algebraic,nister2004efficient, kukelova2008automatic} within a RANSAC framework~\cite{fischler1981random, chum2003locally, raguram2012usac}.
These problems estimate model parameters from the minimal number of point correspondences and in most cases reduce to solving systems of polynomial equations.

Many camera pose estimation and calibration problems can be formulated as minimal problems.
A classical example is the P3P (Perspective-Three-Point) problem, which estimates the camera pose (rotation and translation) from three 3D-2D point correspondences~\cite{haralick1991analysis}.
Several recent methods address the P3P problem by formulating it as finding the intersection of two conics~\cite{wu2025conic,Ding_2023_CVPR,larsson2018beyond}.
Another classical example is the five-point problem, which estimates the relative pose between two calibrated views from five point correspondences~\cite{nister2004efficient}. 
The relative pose problem has been extended to include additional unknown parameters, such as focal length and radial distortion~\cite{kukelova2007minimal, stewenius2008minimal, jiang2014minimal}.

Minimal problems often result in complex systems of multivariate polynomial equations in several variables.
A widely used approach for solving such systems in computer vision is based on Gröbner basis method~\cite{byrod2007improving, byrod2009fast, kukelova2013algebraic, kukelova2008automatic, larsson2017efficient, larsson2018beyond}. 
Its use was popularized by Stewénius~\cite{stewenius2005grobner}, and early solvers based on this approach were typically handcrafted~\cite{stewenius2006recent, stewenius2008minimal}, often exhibiting limited numerical stability~\cite{stewenius2005hard}.
Later research focused on automating solver generation~\cite{kukelova2008automatic, larsson2017efficient, larsson2017polynomial}, improving numerical robustness~\cite{byrod2007improving, byrod2008column}, and optimizing runtime performance~\cite{larsson2017efficient, larsson2017polynomial, larsson2018beyond}. 
In particular, Kukelova \etal~\cite{kukelova2008automatic} and Larsson \etal~\cite{larsson2017efficient} developed tools for the automatic generation of efficient and numerically stable Gröbner basis solvers, and Li \etal~\cite{li2020gaps} subsequently introduced GAPS as an improved and more stable extension of this framework.


An alternative approach is based on multipolynomial resultants, which eliminates variables to yield a univariate polynomial in a hidden variable whose roots correspond to the solutions of the original system~\cite{cox1998using}.
Classical resultants, such as the Macaulay resultant, often need most polynomial coefficients to be nonzero, the roots to be distinct, and no solutions to exist at infinity.
However, in practical engineering problems, the equations are often sparse, which limits the applicability of classical resultants. 
To address this limitation, sparse resultants~\cite{emiris1995efficient, Heikkila_2017_ICCV,emiris2002symbolic,Bhayani_2020_CVPR} were developed to exploit the sparsity structure of polynomial systems, enabling effective handling of equations with many zero coefficients.
Compared to classical resultants, sparse resultants often yield more compact resultant matrices.
The determinant of such a matrix defines a univariate polynomial in the hidden variable, from which the solutions can be recovered.
However, computing such determinants symbolically is nontrivial in practice. An $N \times N$ determinant may involve up to $N!$ terms, leading to factorial growth in computational complexity. 
For high-dimensional matrices, such direct symbolic evaluation becomes computationally infeasible and prone to numerical instability.



Beyond Gröbner basis and resultant-based approaches, Kukelova \etal~\cite{kukelova2011polynomial} reformulated systems of polynomial equations as polynomial eigenvalue problems (PEPs), enabling minimal relative pose problems to be solved using linear algebra techniques.
However, this method may lead to unnecessarily high-dimensional vector spaces and introduce spurious roots, resulting in numerical instability when solving sparse polynomial systems with high degrees.
To address this, an improved algorithm based on sparse elimination theory was introduced in~\cite{Heikkila_2017_ICCV}, which selects a more compact monomial basis, yielding more valid solutions and improved numerical stability.
Nevertheless, it may produce rank-deficient elimination matrices, which undermine solver stability.
A further improvement was achieved by Bhayani \etal, who introduced an additional equation of a special form and reformulated the resultant constraint as a regular eigenvalue problem~\cite{Bhayani_2020_CVPR}, and further enhanced its efficiency and stability in~\cite{bhayani2024sparse}.
Building upon this approach, a hidden-variable resultant method was presented in~\cite{bhayani2021computing}, which yields a larger eigenvalue problem but avoids unstable matrix inversions, 
achieving higher numerical stability.

In this paper, we build upon sparse resultants and address the bottlenecks of symbolic determinant expansion from a numerical perspective. 
We propose a novel sampling-based, matrix inversion-free method for constructing efficient and numerically stable minimal solvers.
Our main contributions include:
\begin{itemize}[leftmargin=2em]
    \item We replace symbolic determinant expansion with an inverse fast Fourier transform (IFFT)-based interpolation scheme, further accelerated by a tensor-wise fast Fourier transform (FFT) implementation.
    \item We recover the remaining variables using Cramer’s rule and introduce a greatest common divisor (GCD)-based criterion for robust identification of the rank-1 deficient submatrix under floating-point noise.
    \item The proposed solver achieved higher numerical stability and comparable accuracy and runtime to state-of-the-art methods.
\end{itemize}
\section{Theoretical background and related work}
\label{sec:background}

We formulate the minimal problem as a system of $m$ multivariate polynomial equations
\begin{equation}
f_1(x_1, \ldots, x_n) = 0, \ldots, f_m(x_1, \ldots, x_n) = 0,
\label{eq:original_system}
\end{equation}
in $n$ unknown variables $x_1, \ldots, x_n$. 
These variables are ordered and concatenated to form a vector $\V{x} = \begin{bmatrix} x_1, \ldots ,x_n \end{bmatrix}^\top$~\footnote{The ordering can be chosen arbitrarily without loss of generality.}
, and the monomials $\V{x}^{\boldsymbol{\alpha}_{i,j}}$ are defined by
\begin{equation}
\V{x}^{\boldsymbol{\alpha}_{i,j}} = \prod_{k=1}^n x_k^{\alpha_{i,j,k}},
\label{eq:monomials}
\end{equation}
where $\alpha_{i,j,k} \in \mathbb{N}$ is the order of the variable $x_k$ in the $j$th term of the $i$th polynomial. Thus, each polynomial can be expressed as a linear combination of monomials
\begin{equation}
f_i(\V{x}) = \sum_{\boldsymbol{\alpha}_{i,j} \in \mathbb{N}^n} c_{i,\boldsymbol{\alpha}_{i,j}} \, \V{x}^{\boldsymbol{\alpha}_{i,j}},
\label{eq:each_polynomial}
\end{equation}
where $c_{i,\boldsymbol{\alpha}_{i,j}}$ is the coefficient of the monomial \( \V{x}^{\boldsymbol{\alpha}_{i,j}} \) in the \( i \)th polynomial.

\subsection{Resultant-based approach to polynomial solving}
Resultants provide a classical algebraic approach for solving systems of polynomial equations. 
A resultant is defined for a system consisting of \( m = n + 1 \) polynomials in \( n \) variables. 
For a system~\eqref{eq:original_system} with \( m = n + 1 \), the resultant is an irreducible polynomial in the coefficients \( c_{i,\boldsymbol{\alpha}_{i,j}} \), denoted by \(Res([c_{i,\boldsymbol{\alpha}_{i,j}}]) \), which vanishes if and only if the system has a non-trivial common solution. 
For a more formal treatment of resultants and their properties, we refer the reader to Cox \etal~\cite{cox1998using}.

A step in a resultant-based method is to expand the input polynomials \( f_1, \ldots, f_{m} \) into a set of linearly independent polynomials that can be represented as a matrix product as
\begin{equation}
\M{M}([c_{i,\boldsymbol{\alpha}_{i,j}}]) \V{u}=0,
\label{eq:matrix_form_polynomial}
\end{equation}
where \( \M{M}([c_{i,\boldsymbol{\alpha}_{i,j}}]) \) is 
square and full rank for generic values of the coefficients \( c_{i,\boldsymbol{\alpha}_{i,j}} \), and \( \V{u} \) is a vector of monomials of \( \V{x}^{\boldsymbol{\alpha}_{i,j}} \). If \( f_1,\ldots f_m \) have a common root, the determinant of \( \M{M}([c_{i,\boldsymbol{\alpha}_{i,j}}]) \) must vanish. According to the definition of resultants, \( Res([c_{i,\boldsymbol{\alpha}_{i,j}}]) = 0 \) implies that the polynomials  \( f_1,\ldots f_m \) have a common root~\cite{cox1998using}. These two statements imply that, 
\begin{equation}
Res([c_{i,\boldsymbol{\alpha}_{i,j}}]) = 0 \Longrightarrow \det \M{M}([c_{i,\boldsymbol{\alpha}_{i,j}}]) = 0.
\label{eq:hidden_system}
\end{equation}
Thus, the condition \( \det \M{M}([c_{i,\boldsymbol{\alpha}_{i,j}}]) = 0 \) can be used as a necessary constraint on the coefficients \( c_{i,\boldsymbol{\alpha}_{i,j}} \) for which the system \( f_1, \ldots, f_m \) admits a non-trivial solution.

As resultants are defined for a system of one more polynomial than the number of variables,  we can employ them for solving a system of \( n \) polynomials in \( n \) variables.
One way is to hide one of the \( n \) original variables in the coefficient field and treat it as a constant. Another way is to add an additional polynomial together with a new variable and then hide this variable in the coefficient field.
In both cases, the resulting system contains one more equation than the number of variables.
The \( u \)-resultant~\cite{cox1998using} approach adopts the latter strategy by introducing an additional polynomial into the original system,
thereby transforming it into a form suitable for constructing a square resultant matrix.
Inspired by the \( u \)-resultant, Bhayani \etal~\cite{Bhayani_2020_CVPR} proposed a special form of the extra polynomial \( x_i - \lambda \), where \( \lambda \) is the new variable and \(x_i\) is one of the original \(n\) variables. The resultant \( Res([c_{i,\boldsymbol{\alpha}_{i,j}}], \lambda) \) is computed then with \( \lambda \) treated as a constant, producing a matrix \( \M{M}([c_{i,\boldsymbol{\alpha}_{i,j}}], \lambda) \) that is linear in \( \lambda \). 
By selecting an appropriate submatrix, its Schur complement can be used to form a compact eigenvalue problem whose eigenvectors correspond to the solutions for \( x_1, \ldots, x_n \).

The most common alternative approach for solving systems with \( m = n \) (see~\eqref{eq:original_system}) is to hide one of the \( n \) variables by treating it as a constant.  
Without loss of generality, we assume that the hidden variable is \( x_1 \). Then,  \( x_1 \) is regarded as part of the coefficient field in the original system~\eqref{eq:original_system}, leading to a new system of polynomial equations in the remaining variables:
\begin{equation}
g_1(x_2, \ldots, x_n) = 0, \ldots, g_m(x_2, \ldots, x_n) = 0.
\label{eq:hidden_system}
\end{equation}
This yields \( m = n \) polynomials in \( n - 1 \) variables, to which the resultant-based approach can be applied. The resultant, denoted as 
$Res([c_{i,\boldsymbol{\alpha}_{i,j}}], x_1)$, is a polynomial in both the coefficients \( c_{i,\boldsymbol{\alpha}_{i,j}} \) and the hidden variable \( x_1 \).
Algorithms based on hiding a variable expand
the system \( g_1, \ldots, g_m \) 
into a linearly independent set of polynomials, which can be represented in a matrix form as
\begin{equation}
\M{M}([c_{i,\boldsymbol{\alpha}_{i,j}}], x_1)\,\V{b} = 0,
\label{eq:matrix_hidden_form_polynomial}
\end{equation}
where \( \M{M}([c_{i,\boldsymbol{\alpha}_{i,j}}], x_1) \) is a square matrix whose entries are polynomials in \( c_{i,\boldsymbol{\alpha}_{i,j}} \) and \( x_1 \), and \( \V{b} \) is a vector of monomials in \( x_2, \ldots, x_n \). For simplicity, we denote \( \M{M}([c_{i,\boldsymbol{\alpha}_{i,j}}], x_1) \) by \( \M{M}(x_1) \).
This resultant obtained in this way, referred to as the \textit{hidden variable resultant}, is a univariate polynomial in \( x_1 \) , whose roots are \( x_1 \)-coordinates of the common solutions to the original system. For theoretical details and proofs, we refer the reader to Cox \etal~\cite{cox1998using}.
Kukelova \etal~\cite{kukelova2011polynomial, kukelova2013algebraic} and Hartley \etal~\cite{hartley2012efficient} applied the hidden variable approach to relative pose problems, such as the 5-point problem for calibrated cameras and the 6-point problem with unknown focal length. 
In these problems, the original polynomial systems can be directly rewritten in the form~\eqref{eq:matrix_hidden_form_polynomial} after hiding one variable.
For more complex minimal problems, the hidden-variable resultant method often requires generating an extended set of linearly independent polynomials to construct a valid matrix. 
Approaches based on the Dixon resultant~\cite{kasten2019resultant} and modified Macaulay methods~\cite{kukelova2013algebraic} for this expansion tend to be computationally expensive.

\subsection{Sparse resultants}
In practice, many polynomial systems arising in computer vision are sparse, with a large portion of coefficients equal to zero. 
For such systems, it is possible to obtain more compact resultants using specialized algorithms. These resultants are commonly referred to as sparse resultants.
Emiris~\etal~\cite{emiris1993practical, canny2000subdivision} proposed a general approach for constructing sparse resultants using mixed subdivisions of Newton polytopes. 
This method was applied to the 5-point relative pose problem in~\cite{emiris2012general}, but the resulting solver was not particularly efficient in practice.
Heikkilä~\cite{Heikkila_2017_ICCV} later proposed an improved algorithm based on sparse elimination theory, which selects a more compact monomial basis. 
This method transforms~\eqref{eq:matrix_hidden_form_polynomial} to a generalized eigenvalue problem (GEP) and solves for eigenvalues and eigenvectors to compute solutions to unknowns.
However, the method may still yield a rank-deficient matrix \( \M{M}(x_1) \), compromising solver stability.
Moreover, both~\cite{emiris2012general} and~\cite{Heikkila_2017_ICCV} require the input system to contain as many polynomials as unknowns to compute a valid resultant, which limits their applicability since many minimal problems in computer vision involve more equations than unknowns.
Bhayani~\etal~\cite{Bhayani_2020_CVPR} further improved the sparse resultant method by introducing an additional equation of a special form and computing the Schur complement of a selected submatrix of the resultant matrix, thereby transforming the resultant constraint into a compact eigenvalue problem.
They also extended the approach to support systems where the number of equations is greater than or equal to the number of unknowns.

However, solvers based on resultant-based methods~\cite{Bhayani_2020_CVPR} or constructed using Gröbner-basis methods~\cite{larsson2018beyond, li2020gaps}  typically involve a step of explicit matrix inversion.
This step is computationally expensive and often unstable when the matrix is close to singular or ill-conditioned, as observed in noisy or degenerate configurations of minimal problems. 
In practice, the inversion often causes numerical errors, resulting in inaccurate eigenvalues or failed solver generation.
While the method of Bhayani~\etal~\cite{bhayani2021computing} also avoids explicit matrix inversion, it typically leads to a larger eigenvalue formulation and higher computational cost. 
Motivated by this bottleneck, our approach eliminates the need for explicit matrix inversion by directly reconstructing the determinant polynomial from sampled evaluations via an efficient FFT-based interpolation strategy, as detailed in Section~\ref{sec:method}.

\section{Proposed method}\label{sec:method}
\begin{figure*}[h!]
    \centering
    \begin{subfigure}[b]{0.8\textwidth}
        \centering        
        \includegraphics[width=\textwidth]{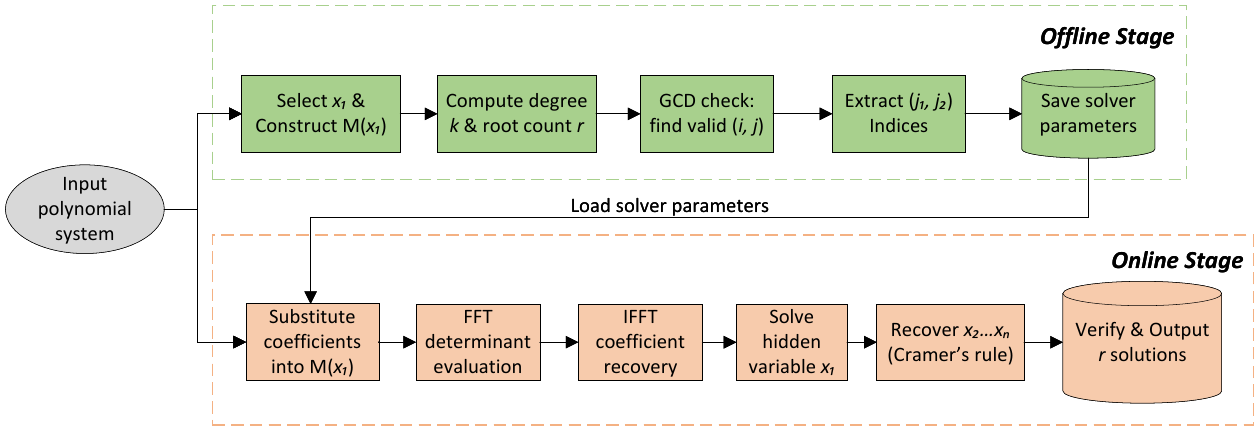}
        \label{fig:pipeline}
    \end{subfigure}
    \caption{High-level overview of the proposed solver pipeline, illustrating the offline construction and online numerical stages.
    }
    \label{fig:pipeline}
\end{figure*}
We begin by constructing a matrix with the same form as in~\eqref{eq:matrix_hidden_form_polynomial} by hiding one of the variables, say \(x_1\), using the hidden variable resultant-based method. Next, we compute the coefficients of the determinant of \(\M{M}([c_{i,\boldsymbol{\alpha}_{i,j}}], x_1)\) as a univariate polynomial in \(x_1\).   
Our approach avoids performing matrix inversion or symbolic expansion for this purpose. Instead, we reconstruct the determinant coefficients numerically from sampled evaluations using IFFT-based interpolation, thereby obtaining a univariate polynomial in the hidden variable.
Solving this polynomial yields the solution to the hidden variable. Subsequently, we can recover the solutions to the remaining variables by identifying a rank-1 deficient submatrix and computing monomial ratios using Cramer’s rule.
To ensure robustness against floating-point noise, we introduce a GCD-based criterion to reliably identify the correct submatrix. The overall pipeline of the proposed solver, including both the offline construction and online recovery stages, is illustrated in~\cref{fig:pipeline}.



\subsection{Hidden variable resultant construction}
We adopt the hidden variable approach to transform the original polynomial system~\eqref{eq:original_system} into the form~\eqref{eq:hidden_system}, where \( x_1 \) is treated as the hidden variable.  Specifically, we consider systems~\eqref{eq:original_system} with at least as many equations as unknowns, i.e., $m \geq n$, which generically admit a finite number of solutions. Our objective is to compute all such solutions.

While Bhayani’s method~\cite{Bhayani_2020_CVPR} constructs sparse resultants 
using an extra variable, we adapt the underlying sparse resultant matrix 
construction technique to our hidden-variable formulation. 
We apply this construction to the hidden-variable system~\eqref{eq:hidden_system}, yielding
\begin{equation}
\M{M}(x_1)\V{b} = \V{0},
\label{eq:matrix_system}
\end{equation}
where \(\M{M}(x_1)\) is the resulting \(N\times N\) resultant matrix.
Here, \( N \) denotes the size of the matrix, determined by the chosen monomial basis.
The determinant of $\M{M}(x_1)$ is a univariate polynomial in $x_1$, 
which can be expressed as
\begin{equation}
\det(\M{M}(x_1)) = c_k x_1^k + c_{k-1} x_1^{k-1} + \ldots + c_0 = 0,
\label{eq:univariate_poly}
\end{equation}
whose roots either correspond to the solutions of the original polynomial system~\eqref{eq:original_system} or to spurious roots.
Let $r$ denote the number of solutions (counting multiplicities) to~\eqref{eq:original_system}. This number \(r\) can be determined for a given problem using algebraic tools such as Gröbner bases or computer algebra systems (e.g., Macaulay2 or Maple).
Note that \(r \leq k\).  
When \(r = k\), all roots of~\eqref{eq:univariate_poly} correspond to valid solutions of~\eqref{eq:original_system}. 
When \(r < k\),
some roots are spurious, introduced by an extraneous factor that typically arises during elimination.
Therefore, the determinant of \(\M{M}(x_1)\) can be expressed in a factorized form
\begin{align}
\det(\M{M}(x_1)) =& q(x_1)p(x_1) \\
=& (q_t x_1^t + \ldots + q_1)(p_r x_1^r + \ldots + p_0) \\
=& 0,
\label{eq:factorized_poly}
\end{align}
where $t = k - r$ is the degree of the extraneous factor $q(x_1)$, and $p(x_1)$ is the univariate polynomial whose $r$ roots correspond to the true solutions of~\eqref{eq:original_system}.
In practice, the extraneous factor \(q(x_1)\) introduces spurious roots that must be filtered out by verifying each candidate solution against the original system~\eqref{eq:original_system} through residual evaluation.



\subsection{Solving the hidden variable with FFT}
\subsubsection{IFFT-based coefficient recovery}
The coefficients $c_k, c_{k-1}, ..., c_0$ in~\eqref{eq:univariate_poly} are multivariate polynomials in the original coefficients $c_{i,\boldsymbol{\alpha}_{i,j}}$. Computing these coefficients through symbolic expansion and evaluating them numerically is often infeasible in practice since an $N \times N$ determinant may involve up to $N!$ terms. 
To address this challenge, we reconstruct the coefficients of $\det(\M{M}(x_1))$ numerically from sampled evaluations using IFFT-based interpolation, evaluating the determinant at uniformly spaced points on the complex unit circle to ensure stability.

For a given minimal problem, the polynomial degree $k$ of $\det(\M{M}(x_1))$ is first determined in an offline stage by assigning random numerical values to the symbolic coefficients and computing the determinant once
(e.g., in Maple or MATLAB). The obtained degree $k$ depends only on the problem formulation and remains constant for all subsequent solver instances.
Once the degree $k$ is determined, $\det(\M{M}(x_1))$ is evaluated at $k+1$ uniformly spaced points on the complex unit circle, defined as 
\begin{equation}
x_1^{(j)} = \omega^{-j}, \quad \omega = e^{2\pi i / (k+1)},
\label{eq:fft_points}
\end{equation}
where $j = 0, 1, \ldots, k$. 
At each sampling point $x_1^{(j)}$, the determinant is computed numerically as
\begin{equation}
y^{(j)} = \det(\M{M}(x_1^{(j)})).
\label{eq:det_sampling}
\end{equation}
This yields a set of $k+1$ evaluations $\{y^{(0)}, y^{(1)}, \ldots, y^{(k)}\}$.
Collecting these evaluations into a vector gives
\begin{equation}
\V{y} =
\begin{bmatrix} 
y^{(0)} & y^{(1)} & \cdots & y^{(k)} 
\end{bmatrix}^\top
= \M{V} \V{c},
\label{eq:vandermonde_system}
\end{equation}
where $\V{c} = [c_0, c_1, \ldots, c_k]^\top$ denotes the coefficient vector of the univariate polynomial in~\eqref{eq:univariate_poly} and $\M{V}$ is a Vandermonde matrix defined by
\begin{equation}
\M{V} =
\begin{bmatrix}
1 & x_1^{(0)} & (x_1^{(0)})^2 & \cdots & (x_1^{(0)})^k \\
1 & x_1^{(1)} & (x_1^{(1)})^2 & \cdots & (x_1^{(1)})^k \\
\vdots & \vdots & \vdots & \ddots & \vdots \\
1 & x_1^{(k)} & (x_1^{(k)})^2 & \cdots & (x_1^{(k)})^k
\end{bmatrix}.
\end{equation}
Directly inverting the Vandermonde matrix \( \M{V} \) to recover the coefficient vector \( \V{c} \) is often numerically unstable, as \( \M{V} \) is typically ill-conditioned. To overcome this, we exploit the fact that when the sampling points \( x_1^{(j)} \) are chosen as the \((k + 1)\)-th roots of unity, \( \M{V} \) becomes equivalent to the discrete Fourier transform (DFT) matrix up to scaling~\cite{von2003modern}. Leveraging this correspondence, we recover \( \V{c} \) by applying an IFFT to the sampled values \( \V{y} \), achieving a matrix inversion-free and numerically stable computation.
The IFFT computes each coefficient $c_l$ as
\begin{equation}
c_l = \frac{1}{k+1} \sum_{j=0}^{k} y^{(j)} \, \omega^{jl}, 
\label{eq:ifft}
\end{equation}
where $l=0,1,\ldots,k.$
This IFFT-based reconstruction requires only $O(k \log k)$ operations, reducing the computational complexity compared with both symbolic expansion and traditional Vandermonde interpolation.
Sampling on the unit circle ensures numerical stability, 
because all sampling points have unit modulus, preventing exponential growth or decay in the monomial terms. 
Moreover, computing the coefficients via the IFFT avoids explicit matrix inversion, 
which further enhances numerical robustness.
\subsubsection{Tensor-wise FFT implementation}
To efficiently compute the determinant evaluations $\{y^{(j)}\}$, 
the matrices $\M{M}(x_1^{(j)})$ are evaluated simultaneously using an FFT-based scheme. 
Each entry of the matrix $\M{M}(x_1)$ corresponds to a univariate polynomial in the hidden variable $x_1$. The $(r,c)$-th entry of $\M{M}(x_1)$ is
\begin{equation}
p_{rc}(x_1)=\sum_{l=0}^{d} a_{rc,l}x_1^{l},
\end{equation}
and $\M{M}(x_1)$ can be compactly represented as
\begin{equation}
\M{M}(x_1) = \sum_{l=0}^{d} \M{A}_l x_1^{l},
\label{eq:sx1}
\end{equation}
where $\M{A}_l = [\,a_{rc,l}\,]_{r,c=1}^{N} \in \mathbb{C}^{N\times N}$.
Stacking the coefficient matrices $\{\M{A}_l\}$ along the polynomial-degree axis yields 
a three-dimensional tensor $\M{A}\in\mathbb{C}^{N\times N\times(d+1)}$. 
Evaluating $\M{M}(x_1)$ at all sampled unit-circle points (as defined in~\eqref{eq:fft_points}) 
is then equivalent to applying a one-dimensional FFT along the third dimension of $\M{A}$:
\begin{equation}
\M{M}_{\text{fft}}(:,:,j)=\sum_{l=0}^{d}\M{A}_l e^{-2\pi i jl/(d+1)}.
\end{equation}
where \( j = 0, 1, \ldots, d \), and $\M{M}_{\text{fft}}(:,:,j)$ denotes the tensor slice corresponding to the evaluated matrix $\M{M}(x_1^{(j)})$, from which the determinant values \( y^{(j)} = \det(\M{M}_{\text{fft}}(:,:,j)) \) are computed in a single batched operation.
This tensor-wise FFT formulation replaces sequential polynomial evaluations,
reducing the overall computational cost while maintaining numerical stability.

Once the coefficient vector is recovered, the resulting univariate polynomial in $x_1$ is solved 
numerically using the companion-matrix eigenvalue method (as implemented in MATLAB’s \textit{roots} function) 
to obtain all candidate values of $x_1$, from which the remaining unknowns are recovered as described in the following section.

\subsection{Recovering remaining variables}

\subsubsection{Variable recovery via Cramer’s rule}
After obtaining the hidden variable $x_1$, substituting it back into~\eqref{eq:matrix_system} yields a homogeneous system. Since the system admits a nontrivial solution, the matrix $\M{M}(x_1)$ is rank-deficient by one.
Accordingly, one row $i$ and one column $j$ can be selected from $\M{M}(x_1)$ such that, after removing them, a full-rank submatrix $\M{M}'(x_1)$ is obtained. 
The system can then be reorganized as a reduced linear system that can be solved for the remaining variables:
\begin{equation}
\begin{bmatrix}
\M{M}'(x_1) & \V{m}'_j \\
\V{m}'^{\mathsf{T}}_i & m_{ij}
\end{bmatrix}
\begin{bmatrix}
\V{b}' \\
b_j
\end{bmatrix}
= \V{0},
\label{eq:schur_block}
\end{equation}
where $\V{m}'_j$ denotes the $j$th column of $\M{M}(x_1)$ with its $i$th row removed, 
$\V{m}'^{\mathsf{T}}_i$ denotes the $i$th row with its $j$th column removed, 
$m_{ij}$ is the $(i,j)$th element of $\M{M}(x_1)$, 
$b_j$ is the monomial corresponding to the $j$th column, 
and $\V{b}'$ contains the remaining monomials of $\V{b}$.
This yields a linear system:
\begin{equation}
\M{M}'(x_1) \V{b}' / b_j = -\V{m}'_j.
\label{eq:linear_ratio}
\end{equation}
Because $\M{M}'(x_1)$ is in general full rank, the vector $\V{b}'/b_j$ can be solved accordingly. 
Instead of explicitly inverting $\M{M}'(x_1)$, Cramer's rule is applied to compute each monomial ratio as
\begin{equation}
\frac{b_k}{b_j} = \frac{|\tilde{\M{M}}'_k(x_1)|}{|\M{M}'(x_1)|},
\label{eq:cramer_ratio}
\end{equation}
where $\tilde{\M{M}}'_k(x_1)$ denotes the matrix obtained by replacing the $k$th column of $\M{M}'(x_1)$ with $-\V{m}'_j$.  
This formulation enables direct extraction of the remaining variables $x_2, \ldots, x_n$ from the computed ratios of monomials.  
For example, if $b_1 = x_2^4 x_3$ and $b_2 = x_2^3 x_3$, we get $x_2 = (b_1 / b_j)/(b_2 / b_j)$ , independently of the choice of $b_j$.
Let us generalize this by assuming that $(j_1, j_2)$ is a pair of indices to the elements of $\V{b}'$ giving the ratio for solving the unknown $x_w$. The value of $x_w$ is then obtained as
\begin{equation}
x_w(x_1) = \frac{|\tilde{\M{M}}'_{j_1}(x_1)|}{|\tilde{\M{M}}'_{j_2}(x_1)|}.
\label{eq:back_sub_solution}
\end{equation}
This process is repeated for each root of $x_1$ and the corresponding $x_w$, where $w = 2, \ldots, n$, to recover all solutions of the original system~\eqref{eq:original_system} (Algorithm~\ref{alg:online_stage}).


\subsubsection{Robust identification of rank-deficient submatrices}
Theoretically,
substituting a root $x_1$ into $\M{M}(x_1)$
yields a rank-1 deficient matrix.
In practice, floating-point errors may lead to inaccurate rank estimation, causing the matrix to be mistakenly treated as full rank.
In such cases, all submatrices obtained by deleting a single row and column may also appear full rank. This makes it difficult to identify a valid pair whose corresponding submatrix enables reliable recovery of the remaining variables.

To address this issue, we propose a more robust identification strategy. 
For all possible row-column deletion pairs $(i,j)$  in $\M{M}(x_1)$, we construct the corresponding submatrix $\M{M}'^{(i,j)}(x_1)$ and test whether $\det(\M{M}(x_1))$ and $\det(\M{M}'^{(i,j)}(x_1))$ are symbolically coprime.
Since $x_1$ is a root of $\det(\M{M}(x_1)) = 0$ but not of $\det(\M{M}'^{(i,j)}(x_1)) = 0$, the two polynomials should be coprime, and their GCD should be a constant:
\begin{equation}
\gcd\left(\det(\M{M}(x_1)), \det(\M{M}'(x_1))\right) = c,
\label{eq:gcd_constant_check}
\end{equation}
where \( c \in \mathbb{R} \setminus \{0\} \).
However, direct symbolic expansion of the determinant, which depends on the symbolic coefficients and the hidden variable $x_1$, is computationally prohibitive for large matrices.
Instead, we adopt a random specialization strategy. 
We assign random prime values to all symbolic coefficients, while keeping $x_1$ symbolic.
We then compute the determinant of $\M{M}(x_1)$ and each candidate $\M{M}'^{(i,j)}(x_1)$, evaluate their GCD as in~\eqref{eq:gcd_constant_check}. Since only $x_1$ remains symbolic in $\M{M}(x_1)$, the determinant computation becomes considerably faster than full symbolic expansion.
The GCD obtained after specialization is equivalent to the symbolic GCD and thus reliably indicates symbolic coprimality.

In our experiments, this approach consistently identified valid row-column pairs across all problems, demonstrating its robustness. 
Hence, if the GCD between $\det(\M{M}(x_1))$ and $\det(\M{M}'^{(i,j)}(x_1))$ is a constant, the pair $(i,j)$ is accepted as a valid deletion yielding a submatrix of rank $(N-1)$. 
This process is performed once in the offline stage and reused for all instances of the same minimal problem, improving solver stability and generalizability (see Algorithm~S1 in the Supplementary Material).

\begin{algorithm}
\caption{Online Stage}
\label{alg:online_stage}

\textbf{Input:} Coefficients $\{c_{i,\boldsymbol{\alpha}_{i,j}}\}$ of the polynomial system~\eqref{eq:original_system}; outputs from the offline stage (see Algorithm~S1 in the Supplementary
Material)\\
\textbf{Output:} Valid solutions $\{\V{x}\}$ to the system
\begin{algorithmic}[1]
\State Substitute the coefficient values into $\M{M}(x_1)$.
\State Generate $d{+}1$ sampling points $x_1^{(j)}$ as in~\eqref{eq:fft_points}.
\State Evaluate all $\M{M}(x_1^{(j)\M{M}})$ using FFT along the degree axis~\eqref{eq:sx1}, and compute $\det(\M{M}(x_1^{(j)}))$ as in~\eqref{eq:det_sampling}.
\State Compute coefficients of $\det(\M{M}(x_1))$ via IFFT.
\State Form $p(x_1)$ and solve $p(x_1){=}0$ to obtain all candidate roots of $x_1$.
\For{each root $x_1^{(i)}$}
    \For{each $x_j \in \{x_2, \dots, x_n\}$}
        \State Find $(j_1, j_2)$ associated with $x_j$.
        \State Substitute $x_1^{(i)}$ and $(j_1, j_2)$ into~\eqref{eq:back_sub_solution} to compute  \phantom{111111}$x_j^{(i)}$.
    \EndFor
\EndFor
\State For cases with extra candidate roots, select the $r$ valid solutions.
\end{algorithmic}
\end{algorithm}

\section{Experiments}

\begin{table*}[t]
\centering
\caption{Numerical stability comparison among our solver, SparseR, and GAPS across different minimal problems.}
\begin{tabular}{clccccccccc}
\toprule
\# & 
\text{Problem} &
\multicolumn{3}{c}{\text{Our}} &
\multicolumn{3}{c}{\text{SparseR~\cite{Bhayani_2020_CVPR}}}&
\multicolumn{3}{c}{\text{GAPS~\cite{li2020gaps}}}\\
\cmidrule(lr){3-5}
\cmidrule(lr){6-8} 
\cmidrule(lr){9-11} 
 & & mean & med. & fail(\%)  & mean & med. & fail(\%) & mean & med. & fail(\%) \\

\midrule
1 & Rel. pose F+$\lambda$ 8pt &-13.17&-13.50&\textbf{0}&-12.71&-13.29&0.2 &-12.62&-13.21&0.1\\
2 & Rel. pose E+$f$ 6pt  &-12.07&-12.42&\textbf{0}&-12.56&-12.80&\textbf{0} &-12.48&-12.84&\textbf{0}\\
3 & Rel. pose $f$+E+$f$ 6pt &-11.41&-11.87&\textbf{0}&-11.40&-11.86&\textbf{0} &-11.05&-11.64&\textbf{0}\\
4 & Stitching $f\lambda$+R+$f\lambda$ 3pt  &-12.92&-13.15&\textbf{0}&-12.73&-13.05& \textbf{0}&-12.74&-13.05&\textbf{0}\\
5 & Rel. pose E+$f\lambda$ 7pt (elim. $\lambda$) &-10.99&-11.58&\textbf{0}&-11.41&-11.77& \textbf{0}&-11.25&-11.63&\textbf{0}\\
6 & Triangulation from satellite im. &-11.88&-12.17&\textbf{0}&-11.64&-11.94&0.02&-11.50&-11.84&0.02 \\
7 & Optimal PnP (Hesch)  &-11.16&-11.46&\textbf{0}&-11.09&-11.39&0.04 &-11.35&-11.71&0.04\\
8 & Rolling shutter pose &-11.14&-11.55&\textbf{0}&-12.43&-12.64&\textbf{0} &-12.26&-12.46&\textbf{0}\\
9 & Abs. Pose P4Pfr (elim. $f$) &-11.19&-11.59&\textbf{0}&-11.26&-11.55&0.04&-11.67&-11.93&\textbf{0} \\
10 & Rel. pose E+$f\lambda$ 7pt &-7.83&-9.98&\textbf{0}&-9.75&-10.13&0.02&-7.65&-8.07&0.3 \\
11 & Abs. pose refractive P5P &-11.24&-11.61&\textbf{0}&-10.68&-11.07&0.06&-9.96&-10.38&0.18 \\
12 & Abs. pose quivers &-9.68&-10.04&\textbf{0}&-9.54&-9.83&0.04 &-9.03&-9.47&0.38\\
13 & Optimal PnP (Cayley) &-8.33&-8.76&\textbf{0}&-8.76&-9.13&0.12 &-8.78&-9.14&0.08\\
14 & Rel. pose $\lambda_1$+F+$\lambda_2$ 9pt &-8.83&-9.42&\textbf{0}&-8.76&-9.33&1.24 &-8.66&-9.22&1.04\\

\bottomrule
\end{tabular}
\label{tab:our_error_results}
\end{table*}

We evaluate the performance of the proposed solver in terms of numerical stability and computational efficiency, and compare it with two state-of-the-art approaches: the sparse resultant-based solver by Bhayani \textit{et al.}~\cite{Bhayani_2020_CVPR} (referred to as \text{SparseR}) and the Gröbner basis solver generator GAPS by Li \textit{et al.}~\cite{li2020gaps}.
The evaluation covers a diverse set of minimal problems in computer vision, including both relative and absolute pose estimation, which enable the verification of the proposed solver's generality and robustness. To make a fair comparison, all solvers are implemented in MATLAB and tested on an Intel Core i5-13500H (2.6\,GHz) CPU with 16\,GB RAM.

\subsection{Numerical stability}
All experiments in this section and Section~\ref{sec:runtime} are conducted on the same synthetic dataset generated following the procedure in~\cite{Bhayani_2020_CVPR}.
Specifically, 5K random instances of data points are created to evaluate and compare the stability and runtime performance of our solver, the SparseR solver, and the solvers generated using GAPS.


The stability measures include the mean and median of $\log_{10}$ of the normalized equation residuals for all computed solutions, as well as the percentage of solver failures, as presented in \cref{tab:our_error_results}. 
For each candidate solution $\mathbf{x}$, we substitute it back into the original system of polynomial equations~\eqref{eq:original_system}, and compute the absolute residual for each equation as $|f_i(\mathbf{x})|$. 
The largest residual among all equations is taken as the error of that solution, denoted as $r(\mathbf{x}) = \max|f_i(\mathbf{x})|.$
To account for the scale of the solution vector, this error is further normalized by the Euclidean norm of $\mathbf{x}$, resulting in the normalized residual $\hat{r}(\mathbf{x}) = r(\mathbf{x}) / \|\mathbf{x}\|_2$, as defined in~\cite{Bhayani_2020_CVPR}.
A smaller $\hat{r}(\mathbf{x})$ indicates a solution that more accurately satisfies the polynomial system. 
Solver failures are defined as instances in which all computed solutions have normalized residuals greater than $10^{-3}$. 
The failure rate (fail\textit{\%}) represents the proportion of such instances among all 5K trials. 
As shown in \cref{tab:our_error_results}, across all minimal problems, our proposed sampling-based, matrix inversion-free solvers exhibits no failures, whereas SparseR and GAPS solvers show a small percentage of failures in several cases. 
These failures are typically below $0.2\%$, which indicates occasional numerical instability due to floating-point arithmetic.
Furthermore, the mean and median $\log_{10}$ residuals achieved by our method are consistently comparable to, or slightly better than, those of SparseR and GAPS, confirming that the proposed determinant computation maintains high numerical precision without matrix inversion.

\begin{table*}[t]
\centering
\captionsetup{width=1\linewidth}
\caption{Runtime comparison among our solver, SparseR, and GAPS across different minimal problems. 
Values marked with $^\ast$ denote results before filtering spurious roots. 
}
\begin{tabular}{clcccccc}
\toprule
\# & 
\text{Problem} &
\multicolumn{2}{c}{\text{Our}} &
\multicolumn{2}{c}{\text{SparseR~\cite{Bhayani_2020_CVPR}}}&
\multicolumn{2}{c}{\text{GAPS~\cite{li2020gaps}}} \\
\cmidrule(lr){3-4}
\cmidrule(lr){5-6} 
\cmidrule(lr){7-8} 
 & & time (ms) &roots & time (ms)&roots& time (ms)&roots\\
\midrule
1 & Rel. pose F+$\lambda$ 8pt (8 sols) &\textbf{0.086}&8&0.101&9&0.106&8\\
2 & Rel. pose E+$f$ 6pt (9 sols) &\textbf{0.154}&9&0.345&9&0.235&9\\
3 & Rel. pose $f$+E+$f$ 6pt (15 sols) &\textbf{0.287}&{15}&0.597&18 &0.515&{15}\\
4 & Stitching $f\lambda$+R+$f\lambda$ 3pt (18 sols) &\textbf{0.213}&{18}&0.232&{18} &0.241&{18}\\
5 & Rel. pose E+$f\lambda$ 7pt (elim. $\lambda$) (19 sols) &\textbf{0.417}&{19}& 0.637&{19}&0.545&{19}\\
6 & Triangulation from satellite im. (27 sols) &$2.923$&{27}&\textbf{0.552}&{27} &0.554&{27}\\
7 & Optimal PnP (Hesch) (27 sols) &$2.773$&{27}&\textbf{0.688}&{27}&1.454&{27}\\
8 & Rolling shutter pose (8 sols) &$1.526^{*}$&11&0.543&{8}&\textbf{0.320}&{8}\\
9 & Abs. Pose P4Pfr (elim. $f$) (12 sols) &$0.773^{*}$&14&\textbf{0.662}&{12} &0.718&{12}\\
10 & Rel. pose E+$f\lambda$ 7pt (19 sols) &$\textbf{1.254}^{*}$&24&1.623&{19} &1.445&{19}\\
11 & Abs. pose refractive P5P (16 sols) &$5.863^{*}$&24&\textbf{1.802}&25 &4.022&{16}\\
12 & Abs. pose quivers (20 sols) &$3.221^{*}$&24&\textbf{0.947}&24&2.996&{20}\\
13 & Optimal PnP (Cayley) (40 sols) &$7.442^{*}$&43&\textbf{1.393}&{40}&3.005&{40}\\
14 & Rel. pose $\lambda_1$+F+$\lambda_2$ 9pt (24 sols) &$9.520^{*}$&36& \textbf{1.102}&27&4.377&{24}\\
\bottomrule
\end{tabular}
\label{tab:time_results}
\end{table*}

\begin{figure*}[h!]
    \centering
    \begin{subfigure}[b]{0.3\textwidth}
        \centering        
        \includegraphics[width=\textwidth]{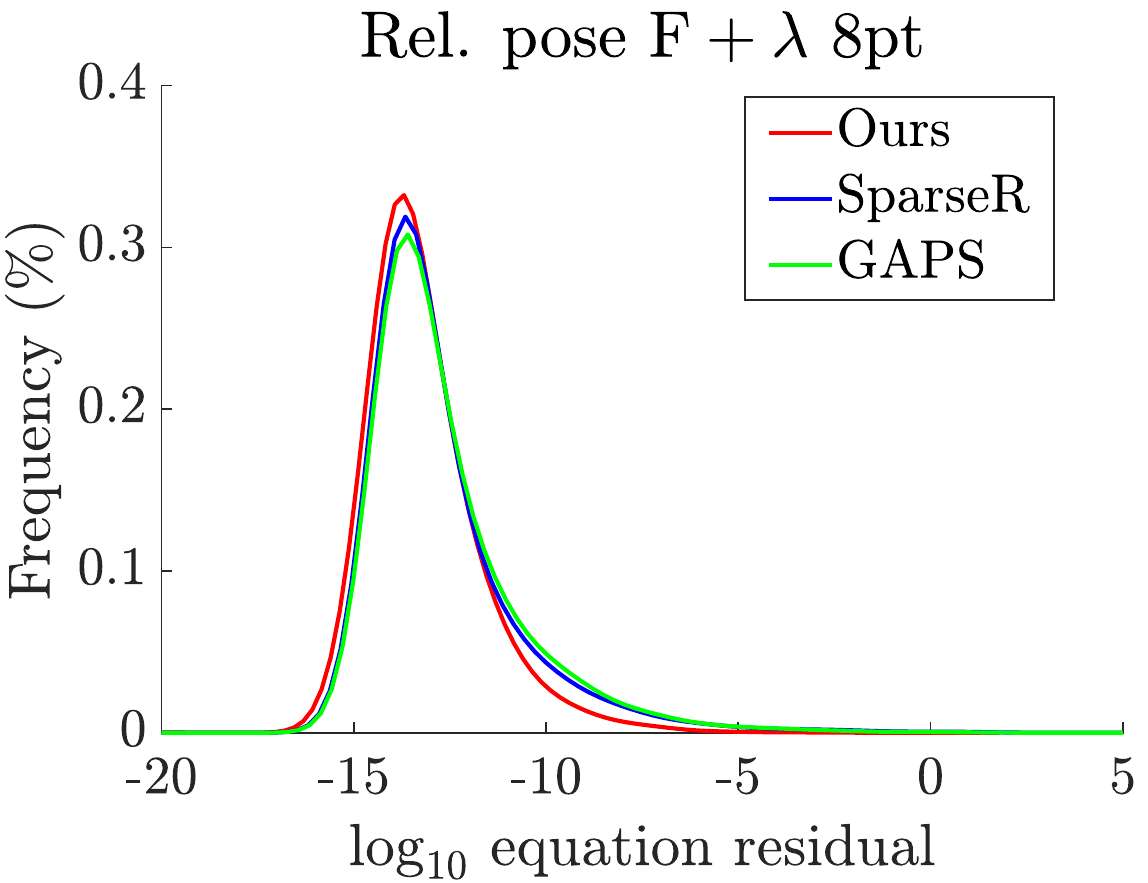}
        \label{fig:hyperbola_points}
    \end{subfigure}
    \hfill
    \begin{subfigure}[b]{0.3\textwidth}
        \centering
        \includegraphics[width=\textwidth]{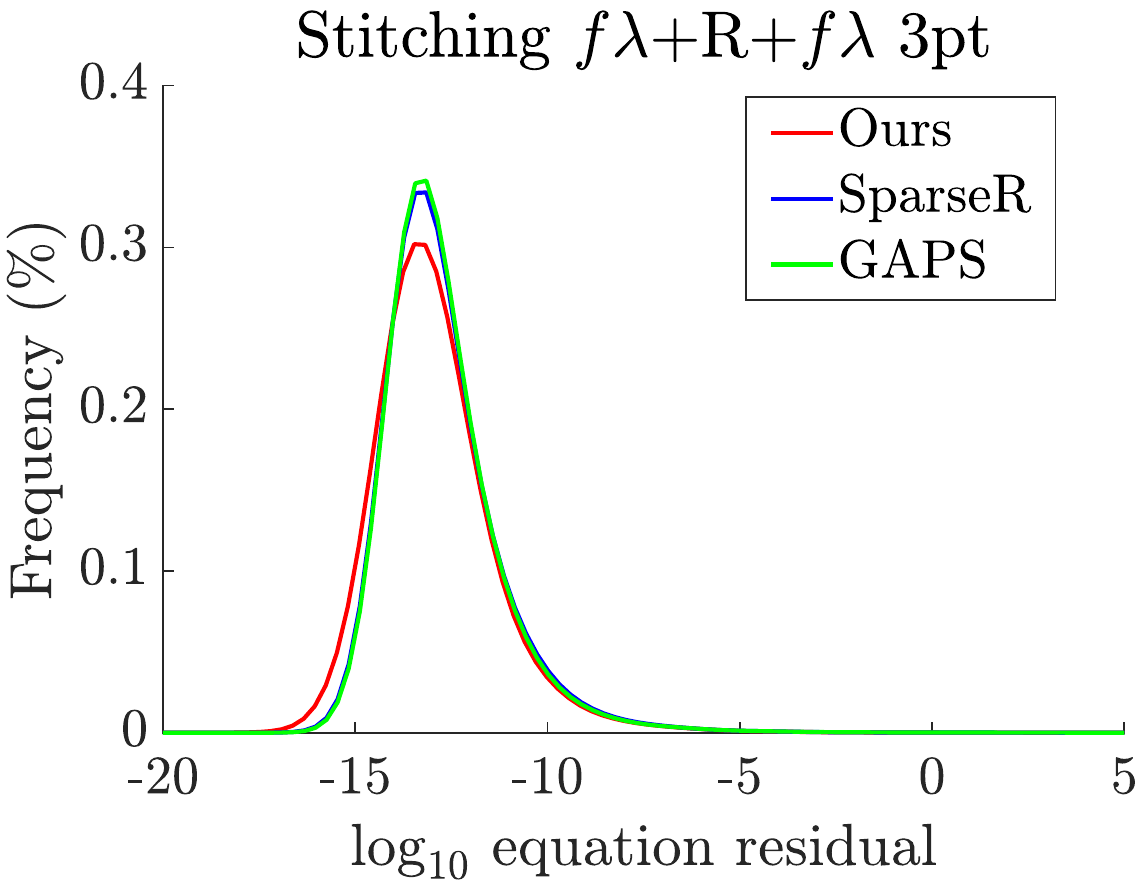}
        \label{fig:ellipse_points}
    \end{subfigure}
    \hfill
    \begin{subfigure}[b]{0.3\textwidth}
        \centering
        \includegraphics[width=\textwidth]{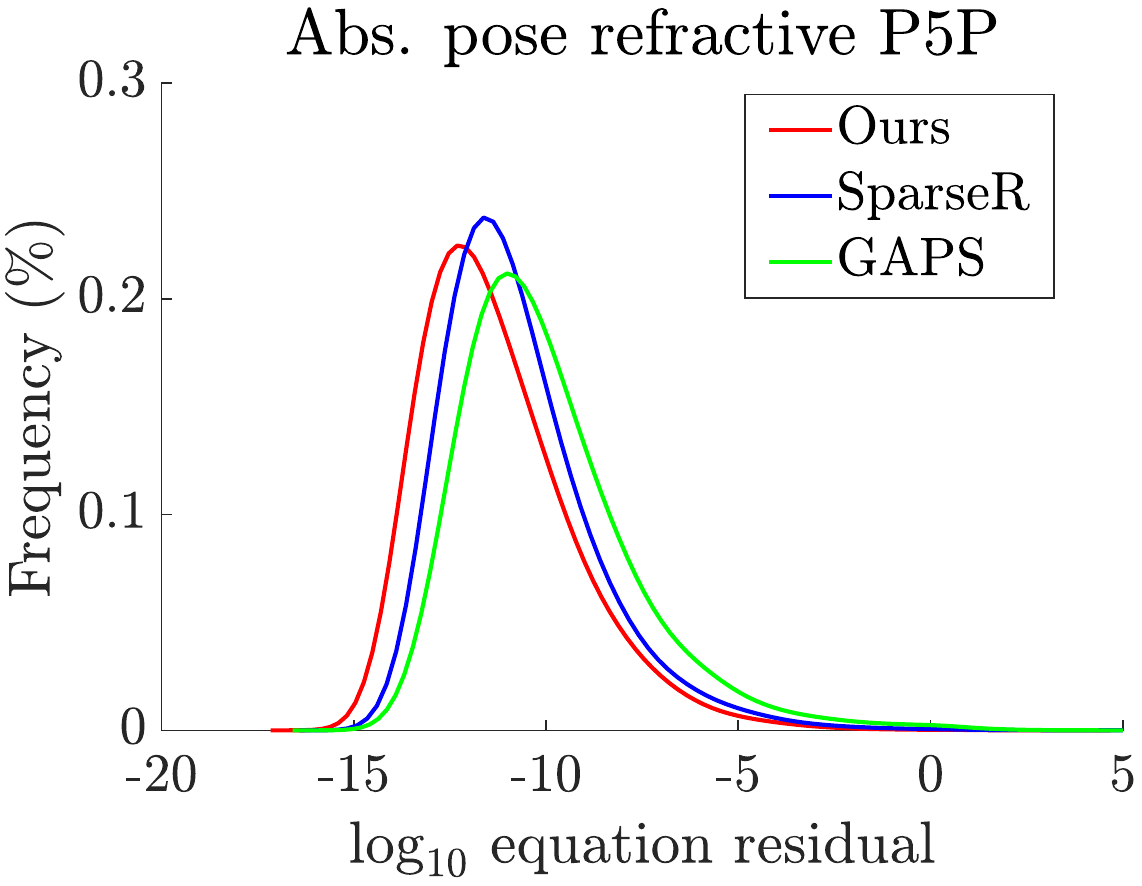}
        \label{fig:parabola_points}
    \end{subfigure}    
    \vspace{-8pt}
    \caption{Histograms of Log10 of normalized equation residual error for three selected minimal problems.}
    \label{fig:residual_hist}
\end{figure*}

As shown in the \textit{roots} column of \cref{tab:time_results}, which reports the number of valid computed solutions, our solver produces the exact theoretical number of valid solutions for the first seven problems.
For the remaining problems
the reconstructed determinant polynomial yields slightly more roots than expected.
Compared to SparseR, our solver yields fewer candidate roots than SparseR in several problems, including the Rel. pose F+$\lambda$ 8pt, Rel. pose $f$+E+$f$ 6pt, and Abs. pose refractive P5P cases, resulting in a cleaner and more compact set of valid solutions.
In our solver, cases with extra candidate roots are verified by back-substitution into the original polynomial equations, and their normalized residuals $\hat{r}(\mathbf{x})$ are computed. 
This residual evaluation serves as a filtering step that removes incorrect or spurious roots arising from numerical errors. 
All candidate roots are then ranked according to their residual magnitudes, and only the top-$r$ solutions with the smallest residuals are retained as the final valid results used for evaluation.
We further visualize the distribution of normalized equation residuals 
for three representative minimal problems from~\cref{tab:our_error_results}. 
As shown in~\cref{fig:residual_hist}, the $\log_{10}$ equation residual histograms 
are consistent with the quantitative results in~\cref{tab:our_error_results}, 
indicating that our proposed solver 
achieves stable performance with low residuals.

\subsection{Runtime Comparison}
\label{sec:runtime}
Tested on the same dataset, our solver is compared with both SparseR~\cite{Bhayani_2020_CVPR} and GAPS~\cite{li2020gaps} in terms of average runtime and the number of computed roots, as presented in~\cref{tab:time_results}, where the \textit{time} column indicates the average runtime per instance.
As shown in table, our solver consistently achieves faster execution for the first five small-scale problems, reducing the runtime by approximately 30\% on average (ranging from 10\% to 50\%) compared to SparseR and GAPS.
It can be observed that the runtime of our solver tends to increase with the number of computed roots. 
This is because our solver evaluates the resultant polynomial for each candidate root of the hidden variable and performs variable recovery repeatedly for all roots.
As the number of roots increases, these repeated evaluations and back-substitutions accumulate, leading to an approximately linear growth in runtime.
For problems with a similar number of roots, the runtime of our solver is mainly influenced by the size of the resultant matrix 
and the complexity of computing its polynomial entries. The corresponding matrix sizes are provided in Supplementary Material. We note that there is a straightforward possibility of parallelizing the implementation using a GPU to improve the speed for large resultant matrices. However, this is left for future work.

\section{Conclusion}
We propose a sampling-based, matrix inversion-free method for constructing efficient minimal solvers.
The method constructs a resultant matrix with respect to a selected hidden variable and numerically reconstructs its determinant coefficients via an IFFT-based strategy, avoiding symbolic expansion and explicit matrix inversion. 
With a robust strategy for identifying rank-deficient submatrices, the proposed solver maintains numerical stability under floating-point noise and enables efficient recovery of all unknowns via Cramer’s rule. 
Experiments on a diverse set of relative and absolute pose estimation problems demonstrate strong numerical stability and competitive runtime performance, particularly on small-scale problems.
These results indicate that the method offers a practical and scalable alternative to traditional Gröbner-basis and resultant-based solvers.
The code is available at: \url{https://github.com/hayden-86/fft-minimal-solvers}.

\section{Acknowledgement}
This work was supported by Research Council of Finland under the project "Methods and Applications for High-Efficiency Polynomial Solvers" (grant no. 355970).
{
    \small
    \bibliographystyle{ieeenat_fullname}
    \bibliography{main}
}


\end{document}